\documentclass[letterpaper, 10 pt, conference]{ieeeconf}

\IEEEoverridecommandlockouts

\PassOptionsToPackage{table}{xcolor} % must come BEFORE \usepackage{xcolor}
\usepackage{xcolor}

\definecolor{cvprblue}{rgb}{0.21,0.49,0.74}
\definecolor{LightGreen}{rgb}{0.88,1,0.88}
\definecolor{LightRed}{rgb}{1,0.88,0.88}
\definecolor{BrightBrownBase}{rgb}{0.8,0.5,0.2}
\colorlet{BrightBrown}{BrightBrownBase!40!white}
\definecolor{verylightgray}{RGB}{245,245,245}

\usepackage{amsmath}
\usepackage{amssymb}
\usepackage{graphicx}
\usepackage{booktabs}
\usepackage{multirow}
\usepackage{array}
\usepackage{float}
\usepackage{tabularx}
\usepackage{colortbl}
\usepackage{makecell}
\usepackage{longtable}
\usepackage{adjustbox}
\usepackage{verbatim}
\usepackage{fancyvrb}

\usepackage[caption=false,font=footnotesize]{subfig}

\usepackage{tikz}
\usetikzlibrary{positioning,shapes.geometric,arrows.meta,calc,patterns}
\usepackage{pgfplots}
\usepgfplotslibrary{polar,groupplots}
\usepackage{pgfplotstable}
\pgfplotsset{compat=1.16}

\usepackage{xspace}
\newcommand{\vlm}{{\textit{V\textsuperscript{2}LM}}\xspace}
\newcommand{\vlms}{{\textit{V\textsuperscript{2}LMs}}\xspace}

\makeatletter
\@ifundefined{subfigcapskip}{}{}
\makeatother

\usepackage[hidelinks]{hyperref}

\usepackage[capitalize]{cleveref}
\crefname{section}{Sec.}{Secs.}
\Crefname{section}{Section}{Sections}
\Crefname{table}{Table}{Tables}
\crefname{table}{Tab.}{Tabs.}
\usepackage[caption=false,font=footnotesize]{subfig}

\title{\LARGE \bf Toward Inherently Robust VLMs Against Visual Perception Attacks}

\author{%
  \protect\parbox{\textwidth}{%
    \centering
    Pedram MohajerAnsari$^{1}$, Amir Salarpour$^{1}$, Michael K{\"u}hr$^{2}$, Siyu Huang$^{1}$, Mohammad Hamad$^{2}$,\linebreak
    Habeeb Olufowobi$^{3}$, Sebastian Steinhorst$^{2}$, Bing Li$^{1}$, Mert D. Pes\'{e}$^{1}$%
  }%
  \thanks{$^{1}$Pedram MohajerAnsari, Amir Salarpour, Siyu Huang, Bing Li, and Mert D. Pesé are with Clemson University, Clemson, SC, USA
    \{\texttt{pmohaje}, \texttt{asalarp}, \texttt{siyuh}, \texttt{bli4}, \texttt{mpese}\}@clemson.edu}
  \thanks{$^{2}$Michael Kühr, Mohammad Hamad, and Sebastian Steinhorst are with the Technical Universität München, Munich, Germany
    \{\texttt{michael.kuehr}, \texttt{mohammad.hamad}, \texttt{sebastian.steinhorst}\}@tum.de}
  \thanks{$^{3}$Habeeb Olufowobi is with the University of Texas at Arlington, Arlington, TX, USA
    \texttt{habeeb.olufowobi@uta.edu}}
}

\begin{document}
\maketitle
\thispagestyle{empty}
\pagestyle{empty}

\begin{abstract}
Autonomous vehicles rely on deep neural networks (DNNs) for traffic sign recognition, lane centering, and vehicle detection, yet these models are vulnerable to attacks that induce misclassification and threaten safety. Existing defenses (e.g., adversarial training) often fail to generalize and degrade clean accuracy. We introduce Vehicle Vision–Language Models (\vlms), fine-tuned Vision Language Models (VLMs) specialized for AV perception, and show that they are inherently more robust to unseen attacks without adversarial training, maintaining substantially higher adversarial accuracy than conventional DNNs. We study two deployments: \emph{Solo} (task-specific \vlms) and \emph{Tandem} (a single \vlm\ for all three tasks). Under attacks, DNNs drop 33\%--74\%, whereas \vlms\ decline by under 8\% on average. Tandem achieves comparable robustness to Solo while being more memory-efficient. We also explore integrating \vlms\ in parallel with existing perception stacks to enhance resilience. Our results suggest \vlms\ are a promising path toward secure, robust AV perception. Code and data are available at \textcolor{blue}{\url{https://github.com/pedram-mohajer/V2LM}}.
%\mert{How about data?}
\end{abstract}

% ------------------ Main content ----------------------------------------------
\section{Introduction}
\label{sec:1-introduction}

Autonomous vehicles (AVs) rely on automated driving systems (ADS) consisting of perception, planning, and control. The perception module uses deep neural network (DNN) models for image classification and object detection, processes camera and LiDAR data~\cite{wen2022deep}, and passes the resulting scene understanding to the planning module for navigation decisions, which are then executed by the control module~\cite{schwarting2018planning}. Prior point-cloud analysis methods include plane-kernel convolutions~\cite{peyghambarzadeh2020point} and non-parametric designs using Gaussian positional encoding~\cite{salarpour2024pointgn}. Although these DNN-based perception models recognize, monitor, and predict nearby objects~\cite{zhang2022adversarial}, they are vulnerable to physical attacks that manipulate real-world environments to deceive perception~\cite{jia2022fooling}, causing misclassification and unsafe driving behaviors, as shown by the DRP-Attack~\cite{sato2021dirty} on lane detection. Similar reliability concerns also arise across other applied ML pipelines beyond autonomous driving, including medical imaging and analytics and scientific simulation workflows~\cite{soltani2025explainable,saberian2025hydroquantum,afshin2025breast}.

Various defense methods have been proposed to mitigate such attacks: defensive distillation reduces model sensitivity but struggles to generalize across diverse perturbations~\cite{papernot2016distillation}; input transformations can suppress adversarial noise but often degrade clean data quality, lowering accuracy~\cite{guo2017countering}; and provable defenses, though theoretically robust, are computationally expensive and difficult to scale~\cite{wong2018provable}. Among these, adversarial training is the most widely adopted defense~\cite{madry2017towards}.

\begin{figure}[t]
  \centering

  % PGFPlots defaults for THIS figure
  \pgfplotsset{
    compat=1.16,
    every axis/.style={
      ybar,
      bar width=0.16cm,
      width=0.65\linewidth,      % <-- same number
      height=0.65\linewidth,     % <-- same number
      ymin=0, ymax=1.1,
      axis line style={draw=none},
      ymajorgrids,
      major grid style={draw=gray!40,dashed},
      ytick={0,0.2,0.4,0.6,0.8,1.0},
      xticklabel style={yshift=-3pt,font=\scriptsize},
      yticklabel style={font=\scriptsize},
      set layers,
      clip=false,
      scale only axis            % keep line widths/fonts unchanged when scaling
    }
  }
  \def\clusterwidth{0.60}   % = 4 bars × 0.15 cm
  \def\halfcluster{0.30}    % = 0.5 × \clusterwidth
  \def\bgheight{1.1}

  % ===== (a) Adversarial =====
  \subfloat[Accuracy on \textbf{adversarial}]{
    \begin{minipage}[t]{0.40\linewidth}
      \centering\scriptsize
      \begin{tikzpicture}
        \begin{axis}[
          xtick={0,1},
          xticklabels={\colorbox{yellow!20}{Before AT}, \colorbox{cyan!10}{After AT}},
          enlarge x limits=0.3,
          ylabel={Accuracy},
          legend style={at={(1.40,1.033)},anchor=south,legend columns=4,font=\scriptsize},
          xtick style={draw=none},
        ]
          % background clusters
          \begin{pgfonlayer}{axis background}
            \path[fill=yellow!20,draw=none]
                 (-\halfcluster,0) rectangle ++(\clusterwidth,\bgheight);
            \path[fill=cyan!10,draw=none]
                 (1-\halfcluster,0) rectangle ++(\clusterwidth,\bgheight);
          \end{pgfonlayer}

          % bars
          \addplot+[fill=blue!40,draw=teal, pattern=north east lines,pattern color=teal]
            coordinates {(0,0.9804) (1,0.9434)};
          \addplot+[fill=blue!40,draw=blue, pattern=dots,pattern color=blue]
            coordinates {(0,0.8293) (1,0.8492)};
          \addplot+[fill=red!60,draw=red, pattern=crosshatch,pattern color=red]
            coordinates {(0,0.5582) (1,0.6899)};
          \addplot+[fill=green!50,draw=black, pattern=horizontal lines,pattern color=black]
            coordinates {(0,0.2312) (1,0.598)};
          \legend{NVILA, Qwen-VL, YOLOv5-cls, ViT-small}
        \end{axis}
      \end{tikzpicture}
    \end{minipage}
  }
  \hfill
  % ===== (b) Benign =====
  \subfloat[Accuracy on \textbf{benign}]{
    \begin{minipage}[t]{0.40\linewidth}
      \centering\scriptsize
      \begin{tikzpicture}
        \begin{axis}[
          xtick={0,1},
          xticklabels={\colorbox{yellow!20}{Before AT}, \colorbox{cyan!10}{After AT}},
          enlarge x limits=0.3,
          ylabel={Accuracy},
          xtick style={draw=none},
        ]
          \begin{pgfonlayer}{axis background}
            \path[fill=yellow!20,draw=none]
                 (-\halfcluster,0) rectangle ++(\clusterwidth,\bgheight);
            \path[fill=cyan!10,draw=none]
                 (1-\halfcluster,0) rectangle ++(\clusterwidth,\bgheight);
          \end{pgfonlayer}

          \addplot+[fill=blue!40,draw=teal, pattern=north east lines,pattern color=teal]
            coordinates {(0,0.9904) (1,0.9916)};
          \addplot+[fill=blue!40,draw=blue, pattern=dots,pattern color=blue]
            coordinates {(0,0.9580) (1,0.9352)};
          \addplot+[fill=red!60,draw=red, pattern=crosshatch,pattern color=red]
            coordinates {(0,0.9488) (1,0.7398)};
          \addplot+[fill=green!50,draw=black, pattern=horizontal lines,pattern color=black]
            coordinates {(0,0.962) (1,0.684)};
        \end{axis}
      \end{tikzpicture}
    \end{minipage}
  }

  \caption{Impact of adversarial training (AT) on NVILA, Qwen-VL, YOLOv5-cls, and ViT-small. (a) Adversarial inputs. (b) Benign inputs.}
  \label{fig:adv_comparison_merged}
\end{figure}

To address these limitations and achieve robust and generalizable models for AV perception tasks, this work proposes a novel finding:~\emph{VLMs inherently exhibit robustness against unseen adversarial attacks on AV perception systems even without adversarial training, significantly outperforming DNNs.} VLMs such as \texttt{Qwen-VL}~\cite{bai2023qwenvl} and \texttt{NVILA}~\cite{liu2024nvila} demonstrate strong resilience against adversarial inputs without adversarial training. Based on this observation, we are the first to systematically fine-tune VLMs for AV perception tasks; fine-tuning trains these VLMs on TSR, ALC, and VD, enabling a fair comparison with task-specific DNNs, and we name them \textbf{V}ehicle \textbf{V}ision \textbf{L}anguage \textbf{M}odels (\vlms).

Thus, we benchmark both conventional DNNs and \vlms \emph{before and after} adversarial training (AT) on clean and adversarial inputs under \textit{\textbf{unseen}} adversarial attacks (``unseen'' means these attacks are never used during training or fine-tuning of the DNNs or VLMs and appear only at test time). As shown in \autoref{fig:adv_comparison_merged}, this comparison highlights two findings. \textbf{(i) VLMs exhibit robustness:} before AT, \texttt{NVILA-8B} remains near 98\% and \texttt{Qwen-VL} about 83\% on adversarial inputs, while both stay above 95\% on benign inputs; after AT, these numbers change by no more than $\pm$3 percentage points (pp), underscoring their insensitivity to the usual AT trade-offs~\cite{raghunathan2019adversarial}. \textbf{(ii) Conventional DNNs collapse under attack, and AT provides only modest relief—and at a cost:} \texttt{YOLOv5-cls}~\cite{yolov5} plunges from roughly 95\% to 56\% (-39 pp), and \texttt{ViT-small}~\cite{dosovitskiy2020image} drops from 96\% to 23\% (-73 pp); although AT increases adversarial accuracy to 69\% for \texttt{YOLOv5-cls} and 60\% for \texttt{ViT-small}, their benign accuracies simultaneously sink to 74\% and 68\%, reflecting the clean-data penalty noted in previous work~\cite{tsipras2018robustness}.

\begin{table}[t]
\centering
\caption{Classification accuracy (\%) of YOLOv5-cls and ViT-small on GTSRB without defenses (row \textbf{Base Model}) and accuracy change $\Delta$ (pp) when standard defenses are applied to benign and Shadow images. Green ($\uparrow$) and red ($\downarrow$) entries denote improvement or degradation vs.\ the no-defense \textbf{Base Model}.}
\scriptsize
\setlength{\tabcolsep}{2pt}
\renewcommand{\arraystretch}{1.0}
\begin{tabular}{lcccc}
\toprule
& \multicolumn{2}{c}{\textbf{YOLOv5-cls}} & \multicolumn{2}{c}{\textbf{ViT-small}} \\
\cmidrule(lr){2-3} \cmidrule(lr){4-5}
\textbf{Method} & \textbf{benign} & \textbf{shadow} & \textbf{benign} & \textbf{shadow} \\
\midrule
\textbf{Base Model} & 94.88 & 55.82 & 96.20 & 23.12 \\
\cmidrule(lr){2-5}
\textbf{Label Smoothing (LS)} &
\cellcolor{green!20}{$\uparrow 0.87$} &
\cellcolor{red!20}{$\downarrow 3.71$} &
\cellcolor{red!20}{$\downarrow 2.47$} &
\cellcolor{red!20}{$\downarrow 11.71$} \\
\cmidrule(lr){2-5}
\textbf{Dropout (DO)} &
\cellcolor{red!20}{$\downarrow 3.17$} &
\cellcolor{red!20}{$\downarrow 8.26$} &
\cellcolor{red!20}{$\downarrow 1.85$} &
\cellcolor{red!20}{$\downarrow 10.06$} \\
\cmidrule(lr){2-5}
\textbf{JPEG Compression (JPG)} &
\cellcolor{red!20}{$\downarrow 2.81$} &
\cellcolor{red!20}{$\downarrow 11.79$} &
\cellcolor{red!20}{$\downarrow 1.74$} &
\cellcolor{red!20}{$\downarrow 8.03$} \\
\cmidrule(lr){2-5}
\textbf{Bit-Depth Reduction (BD)} &
\cellcolor{red!20}{$\downarrow 1.75$} &
\cellcolor{red!20}{$\downarrow 0.95$} &
\cellcolor{red!20}{$\downarrow 0.53$} &
\cellcolor{green!20}{$\uparrow 2.27$} \\
\cmidrule(lr){2-5}
\textbf{Random Resized Cropping (RRP)} &
\cellcolor{red!20}{$\downarrow 22.00$} &
\cellcolor{red!20}{$\downarrow 25.42$} &
\cellcolor{red!20}{$\downarrow 4.22$} &
\cellcolor{red!20}{$\downarrow 14.53$} \\
\cmidrule(lr){2-5}
\textbf{Histogram Equalization (HEQ)} &
\cellcolor{green!20}{$\uparrow 1.59$} &
\cellcolor{red!20}{$\downarrow 0.82$} &
\cellcolor{red!20}{$\downarrow 1.03$} &
\cellcolor{green!20}{$\uparrow 1.12$} \\
\bottomrule
\end{tabular}
\label{tab:gtsrb-delta-compact}
\end{table}

We evaluate six VLMs (\texttt{LLaVA-7B}, \texttt{MobileVLM}~\cite{chu2023mobilevlm}, \texttt{LLaVA-13B-LoRA}~\cite{liu2023llava}, \texttt{MoE-LLaVA}~\cite{lin2024moe}, \texttt{Qwen-VL-7B}~\cite{bai2023qwenvl}, and \texttt{NVILA-8B}~\cite{liu2024nvila}) as auxiliary perception modules for traffic-sign recognition (TSR), automated lane centering (ALC), and vehicle detection (VD) tasks. Each model is tested zero-shot and then re-evaluated after task-specific fine-tuning (\textbf{RQ1}). The first four models are LLaMA-based, the last two Qwen-based, chosen for their performance and offline suitability. Task-specific DNN baselines such as \texttt{YOLOv5-cls} and \texttt{ViT-small} for TSR, \texttt{CLRerNet} for ALC, and \texttt{YOLOv5-dt} for VD are fine-tuned on the same data to ensure a fair comparison with the VLMs under both benign and adversarial conditions.

Then, the effectiveness and resilience of both DNN models and \vlms are tested using \textit{unseen} (not used during model training or fine-tuning) adversarial examples (AEs) against three distinct attacks targeting AV perception algorithms (\textbf{RQ2}): (1) Robust and Accurate UV-map-based Camouflage attack (RAUCA) to deceive VD algorithms~\cite{zhou2024rauca}, (2) a physical-world adversarial attack known as the Dirty Road Patch (DRP-Attack) to compromise DNN-based automated lane centering (ALC) models, and (3) shadows cast on traffic signs to attack TSR algorithms~\cite{zhong2022shadows}; these are strong physical attacks evaluated end-to-end and shown to cause safety-relevant failures.

The study then compares two distinct designs for utilizing \vlms and evaluates their performance on the three aforementioned AV tasks and AEs (\textbf{RQ3}). The first design, termed \textit{Solo Mode}, involves separate \vlms, each fine-tuned individually for one of the AV tasks. The second design, named \textit{Tandem Mode}, uses a single \vlm fine-tuned simultaneously for all three AV tasks, aiming to assess whether it can match the robustness of \textit{Solo Mode} across tasks under both benign and adversarial conditions. This paper makes the following contributions:

\begin{itemize}
    \item This work presents a novel finding: fine-tuned VLMs inherently exhibit superior robustness against \textit{\textbf{unseen}} adversarial attacks compared to task-specific DNNs. We highlight a critical limitation of adversarial training: it significantly degrades benign accuracy while providing only limited improvements against adversarial examples. In contrast, VLMs achieve strong adversarial robustness while maintaining high benign accuracy.
    \item We introduce Vehicle Vision Language Models (\vlms) which are fine-tuned VLMs for AV perception tasks: TSR, ALC, and VD. We propose two deployment strategies: \textit{Solo Mode} (separate VLM fine-tuned for each task) and \textit{Tandem Mode} (a single unified \vlm across multiple perception tasks).
    \item We conduct comprehensive experiments to evaluate the robustness of DNN models and \vlms under adversarial conditions. DNN models experience performance drops of 33\%–74\% under attacks, whereas \vlms show reductions of less than 8\% on average, maintaining high adversarial accuracy without additional defense mechanisms.
\end{itemize}

%\mert{Mention that these are strong physical attacks because reviewers might ask why you are not using any of the newer digital attacks. These physical attacks have been evaluated end-to-end to show safety detriment to victim vehicle.}

%\mert{In the following you are basically trying to see if one VLM can do everything or you need separate specialized VLMs. Compare this to the DNN logic in classic AV stack. Do you have one DNN for TSR, one for OD, one for ALC?}

%\mert{The jump from Problem (in APSET) to Solution (next paragraph) is crazy abrupt. You need to add 1-2 sentences here how VLMs are used in AD and if they can defend against adversarial patterns.}

% \mert{This paragraph should come before the previous one, no?} \mert{So pre-trained ones, right?} \mert{What are you trying to accomplish by fine-tuning? To further improve the accuracy? You need to justify this.}

% \mert{You need to change the order. (ii) is the most important observation. Then (iii), then (i). For (i), I dont know if it is important that VLMs barely change after AT. Important lesson here is that even out of the box they are super robust against adversraial inputs and their bengign performance outperforms DNNs as well. Are the numbers in Figure 1 for NVILA and Qwen for pretrained model or fine tuned?}

\section{Related Work}
\label{sec:2-related_work}

\noindent \textbf{Large Language Models in Autonomous Driving.} Recent works have demonstrated the potential of LLMs in the context of AVs, particularly enhancing perception, control, and motion planning tasks. Regarding perception systems, LLMs utilize external APIs to access real-time information sources, including traffic reports, and weather updates, which significantly enrich the vehicle's ability to gain a comprehensive understanding of its environment~\cite{cui2024receive}. Aldeen \textit{et al.}~\cite{aldeeniv} investigate the application of Large Multimodal Models (LMMs) for enhancing the cybersecurity of AVs. Concerning control, LLMs enable the adjustment of control settings according to driver preferences, thereby personalizing the driving experience~\cite{sha2023languagempc}.  

\noindent \textbf{Adversarial Threats and Defenses.}
Adversarial Examples (AEs) were first defined by Szegedy \textit{et al.}~\cite{szegedy2013intriguing} as subtly modified inputs designed to fool DNNs. These minor, often imperceptible alterations can drastically alter a DNN's predictions~\cite{shen2022sok, cao2021invisible}. To counter these vulnerabilities, various defense mechanisms have been proposed. We re-implemented these methods and evaluated them against the shadow attack~\cite{zhong2022shadows}; as summarized in Table~\ref{tab:gtsrb-delta-compact}, they are largely ineffective and often reduce benign accuracy. Label Smoothing (LS)~\cite{szegedy2016rethinking} softens targets and can underfit fine-grained TSR cues. Dropout (DO)~\cite{srivastava2014dropout} removes capacity needed for small text and borders. JPEG Compression (JPG)~\cite{guo2017countering} suppresses high-frequency edges and characters. Bit-Depth Reduction (BD)~\cite{xu2017feature} distorts class-specific hues via quantization. Random Resized Cropping (RRP)~\cite{xie2017mitigating} can crop out discriminative regions and destabilize scale. Histogram Equalization (HEQ)~\cite{zuiderveld1994contrast} over-equalizes, amplifying halos and shifting the distribution.

\section{\vlm~as a Defense Mechanism}
\label{sec:4-defense_design}

\subsection{Autonomous Vehicle Perception}

The perception system in AVs is responsible for interpreting the environment through sensor data, such as images, to enable safe and efficient operation. This system performs essential tasks including TSR, ALC, and and VD which is part of object detection. TSR enables the vehicle to follow road rules by recognizing traffic signs~\cite{triki2023real}, ALC keeps the vehicle centered within its lane by identifying road markings~\cite{dubey2024artificial}, and VD detects and classifies other vehicles~\cite{caraffi2012system}. The perception system (PS) can be represented as a function, shown in \autoref{eq:perception_system}, which takes an input image \( \mathcal{I} \in \mathbb{R}^{H \times W \times 3} \) and outputs the results of TSR, ALC, and VD. Specifically, TSR outputs the detected traffic sign's class and bounding box if a sign is present; ALC provides a classification for the appropriate steering command; and VD returns the bounding box and class of any detected car.

\begin{figure}[t]
    \centering
    \subfloat[Shadow Attack\label{fig:shadow_AE}]{
      \begin{minipage}[b]{0.35\linewidth}\centering
        \includegraphics[width=1.25cm,height=1.25cm]{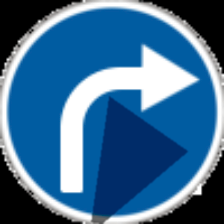}\hspace{2pt}%
        \includegraphics[width=1.25cm,height=1.25cm]{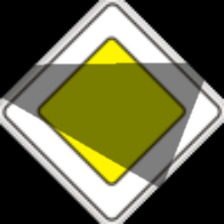}
      \end{minipage}%
    }
    \hfill
    \subfloat[DRP-Attack\label{fig:DRP}]{%
      \begin{minipage}[b]{0.30\linewidth}\centering
        \includegraphics[width=\linewidth,height=1.25cm]{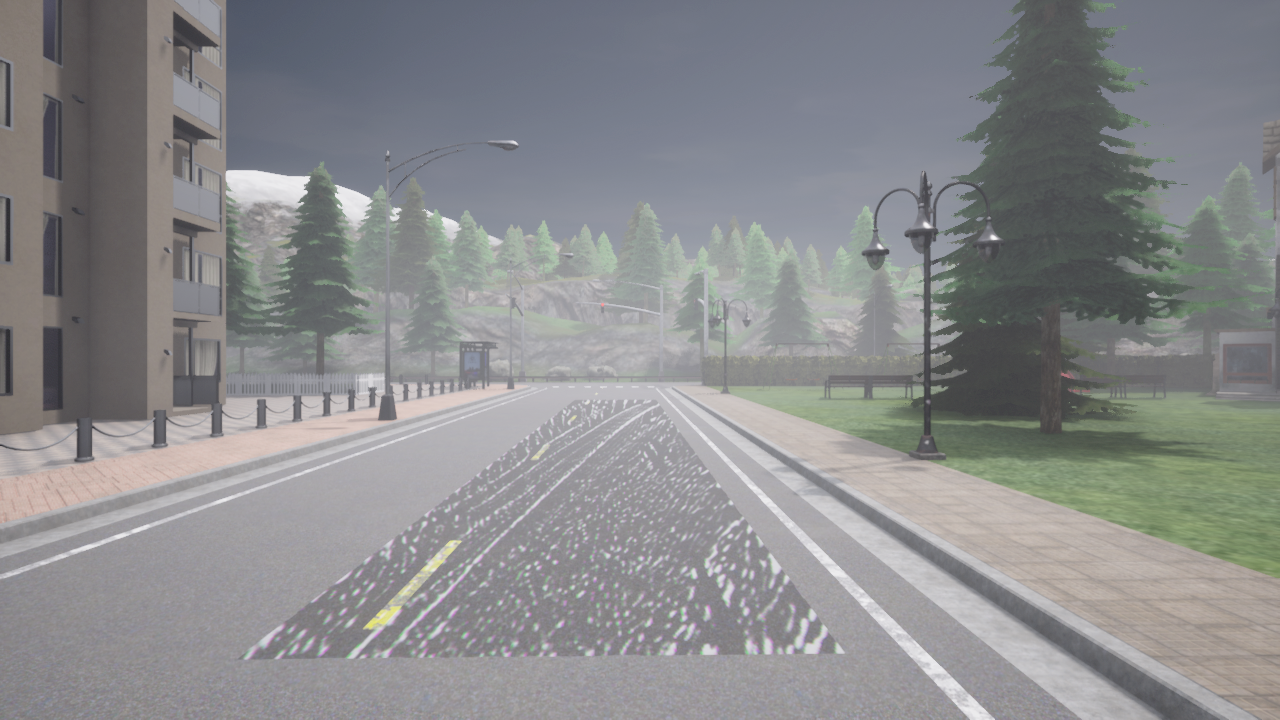}
      \end{minipage}%
    }
    \hfill
    \subfloat[RAUCA\label{fig:RAUCA}]{%
      \begin{minipage}[b]{0.30\linewidth}\centering
        \includegraphics[width=\linewidth,height=1.25cm]{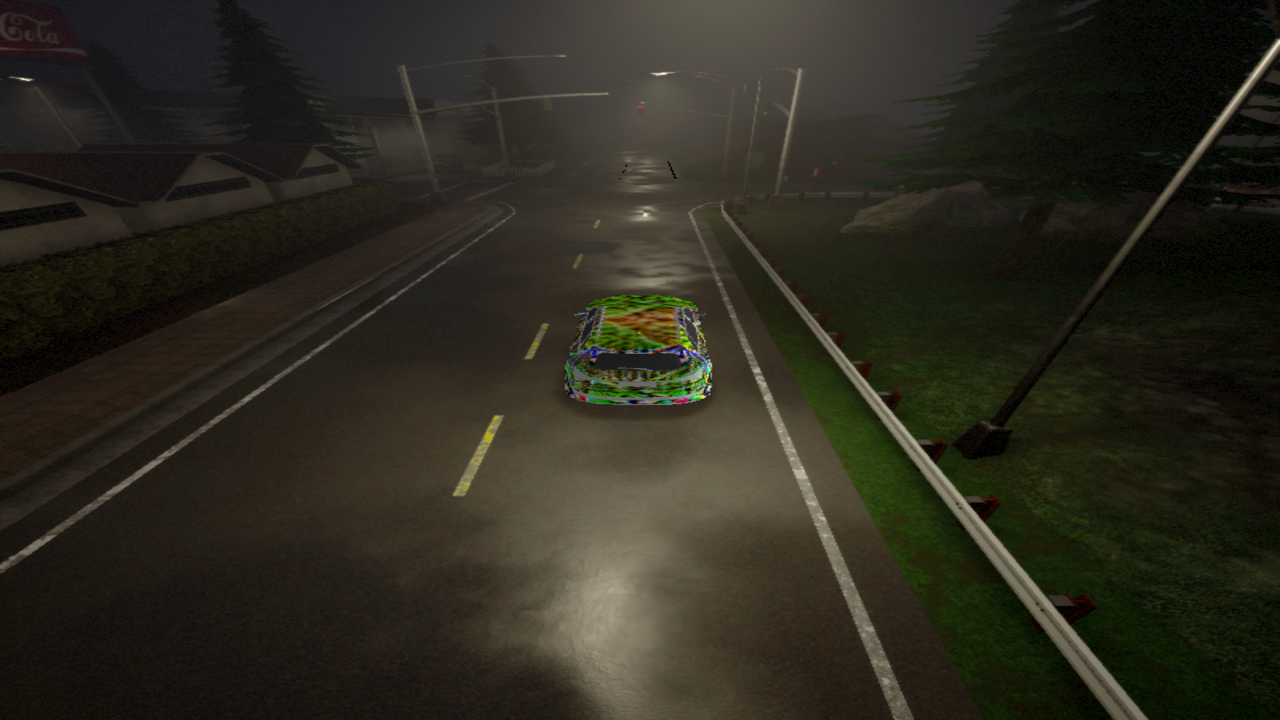}
      \end{minipage}%
    }
    \caption{Examples of adversarial attacks targeting AV perception.}
    \label{fig:combined_images}
\end{figure}

\begin{equation}
\label{eq:perception_system}
\scalebox{0.9}{$
\textit{PS}(\textit{I}) = 
\begin{cases}
   \textit{TSR}(\mathcal{I}) :  \textit{Cls}_{\textit{traffic-sign}} \\
   \textit{ALC}(\mathcal{I}) : \textit{Cls}_{\textit{steering}} \\
   \textit{VD}(\mathcal{I}) : (\textit{Cls}_{\textit{vehicle}}, \textit{BBox}_{\textit{vehicle}})
\end{cases}
$}
\end{equation}

\noindent where \textit{\( \text{Cls} \)} denotes class and \textit{\( \text{BBox} \)} denotes bounding box.

AEs can interfere with the perception system by causing errors in these modules. An AE, \( \mathcal{I}_{\text{adv}} \), is crafted by adding a small perturbation \( \delta \) to an original input \( \mathcal{I} \) such that the target module’s output changes undesirably:

\begin{equation}
\mathcal{I}_{\textit{adv}} = \mathcal{I} + \delta, \quad {\scriptsize \textit{where}} \quad \|\delta\| \leq \epsilon \quad {\scriptsize \textit{and}} \quad f(\mathcal{I}_{\textit{adv}}) \neq f(\mathcal{I})
\end{equation}

\noindent where \( f \) denotes the specific target module in the AV’s perception system, which can be the TSR, ALC, or VD module. This perturbation \( \delta \), constrained by \( \epsilon \), represents a small, controlled modification designed to be imperceptible, thereby ensuring that \( \mathcal{I}_{\text{adv}} \) visually resembles \( \mathcal{I} \). %Despite this, the adversarial perturbation may lead to misclassification of traffic signs, incorrect lane-following commands, or faulty detection of objects.

\subsection{Vision Language Model (VLM)}

\noindent \textbf{Definition.}
VLMs, by integrating visual and textual data~\cite{zhang2024vision}, have the potential to enhance perception tasks by enabling a more comprehensive understanding of the environment. A pre-trained VLM  %\(\text{VLM}_\text{pre}\) 
takes an image and a text prompt as inputs to generate a relevant response: %\mert{I think the following equation can be omitted.}

\begin{equation}
\label{eq:vlm_general}
\textit{VLM}_\textit{pre}(\mathcal{I}, \textit{Prompt}) \rightarrow \textit{Generated-Response}
\end{equation}

This potential arises from their several key strengths: their multimodal learning capability, which allows them to correlate visual and textual information simultaneously~\cite{lu2024theory}; their robustness to variability, which enables them to generalize well across different environments due to extensive training on diverse datasets~\cite{reza2023robust}; their contextual understanding, which leverages textual data to enhance the interpretation of visual scenes~\cite{zhao2024enhancing}; and their comprehensive feature extraction, which combines features from both visual and textual data~\cite{bain2021frozen}. For instance, VLMs could enhance ALC by interpreting road markings and reading associated signs. In VD, VLMs could recognize and classify objects like vehicles by analyzing visual data along with bounding box coordinates. Similarly, VLMs could improve TSR by understanding text on signs, such as speed limits or warnings, to ensure the car follows road rules accurately.

% \begin{equation}\label{eq:solo_eq}
% \textit{Output}_i = \vlm^{i}_{\textit{solo}}(\textit{Data}_i, \textit{Prompt}_i)
% \end{equation}

\noindent \textbf{Fine-Tuning.}
Fine-tuning VLMs on AV-specific tasks is essential for optimizing their performance and enabling adaptation to domain-specific challenges such as variations in lighting, road conditions, and traffic scenarios; the resulting fine-tuned VLM is referred to as \vlm. For the LLaVA family of models, this fine-tuning process involves adjustments to key components. The vision encoder, implemented as CLIP ViT-L/14~\cite{radford2021learning}, extracts visual features from input images. The language model, based on Vicuna (a fine-tuned variant of LLaMA)~\cite{vicuna2023}, interprets textual prompts. A cross-modal attention module integrates the modalities, and all components are jointly optimized via visual instruction tuning~\cite{liu2023visual} to align with autonomous vehicle (AV) tasks.

%\mert{I am trying to understand why you need the following paragraph. Qwen and NVILA are your two base VLMs, but they belong to Experimental Setup/Eval section, not here...}
Qwen-VL adopts a modular architecture comprising an OpenCLIP ViT-bigG visual encoder~\cite{cherti2023reproducible}, a single-layer cross-attention adapter with learnable queries, and a Qwen-7B language model~\cite{bai2023qwen}. The adapter compresses image features to fixed-length sequences, enabling efficient visual grounding and fine-grained perception. Its three-stage training pipeline includes weakly supervised pretraining on large-scale image-text pairs, multi-task visual-language pretraining, and supervised instruction tuning. NVILA builds on this by incorporating a ``scale-then-compress'' design, which first increases spatial and temporal input resolution and then compresses visual tokens for efficient processing. Its architecture consists of a SigLIP-based vision encoder~\cite{zhai2023sigmoid}, a lightweight MLP projector, and a Qwen2-based language model~\cite{yang2024qwen2}. For fine-tuning, NVILA applies lower learning rates to the vision encoder's LayerNorms while optimizing the language model. This allows robust AV adaptation under limited compute budgets~\cite{liu2024nvila}.

%%%%%%%%%%%%%%%%%%%%%%%%%%%%
\begin{figure}[t]
    \centering
    \begin{tikzpicture}[x=1cm,y=1cm,>=Latex]
      \path[use as bounding box] (-1.6,-1.3) rectangle (2.6,1.3);
      \definecolor{boxfill}{RGB}{244,244,244}

      % outer container (dotted frame)
      \node[draw=white, fill=white, dashed, rounded corners,
            minimum width=0.875cm, minimum height=1.65cm] (FRAME) at (0,0) {};

      % two boxes with EXACT same size (reduced vertical gap)
      \tikzset{SameBox/.style={
        draw, rectangle, fill=boxfill, rounded corners,
        text width=2.0cm, minimum height=0.55cm, inner sep=2pt, align=center
      }}
      \node[SameBox] (VLM) at (0, 0.40) {\scriptsize \vlm};
      \node[SameBox] (PS)  at (0,-0.40) {\scriptsize Perception System};

      % input from left → arrows go inside each box
      \coordinate (IN) at (-1.5,0.00);
      \draw (IN) -- ++(-0.30,0) node[left] {\scriptsize I\textsubscript{{\tiny roadview}}};
      \draw[->] (-1.50,0.00) |- (VLM.west);
      \draw[->] (-1.50,0.00) |- (PS.west);

      % outputs from both → merge (moved right) → downward output
      \coordinate (MERGE) at (2,0.00);
      \draw[-] (VLM.east) -- ++(0.60,0) |- (MERGE);
      \draw[-] (PS.east)  -- ++(0.60,0) |- (MERGE);
      \draw[thick, fill=white] (MERGE) circle (0.06);
      \node at (MERGE) {\scriptsize +};
      \draw[->] (MERGE) -- (2.5,0);
      \node[anchor=west] at (2.5,0) {\scriptsize Output\textsubscript{{\tiny all}}};  
    \end{tikzpicture}
    \caption{Possible integration of \vlm~with perception system for attack mitigation. I\textsubscript{roadview} is the image captured by AV cameras, with the output sent to the planning and control modules for action.}
    \label{fig:integration_AV_VLM}
\end{figure}

%#############################

LoRA (Low-Rank Adaptation)~\cite{hu2021lora} offers an efficient alternative by focusing on specific parameters within the model. Only low-rank matrices are introduced in certain layers, primarily within the self-attention and feedforward blocks, allowing the majority of the pre-trained parameters to remain frozen. This technique reduces the memory and computational load while achieving task-specific adjustments by updating only the new, smaller matrices. By leveraging fine-tuning techniques for AV tasks and the efficiency gains of LoRA, exploring \vlms as a solution to AV perception challenges has the potential to address critical limitations of existing methods, enhancing robustness against adversarial attacks while avoiding the performance degradation common in traditional defenses.

% requires: \usepackage[caption=false,font=footnotesize]{subfig}
\begin{figure}[t]
  \centering

  %===================== Solo Mode =====================
  \subfloat[\,Solo Mode\label{fig:Solo_Defense_Design}]{%
    \begin{minipage}[t]{0.40\linewidth}
      \centering
      \begin{tikzpicture}[x=1cm,y=1cm,>=Latex]
        % square canvas
        \path[use as bounding box] (-1.25,-1.25) rectangle (1.25,1.25);
        \definecolor{soloFill}{RGB}{255,255,255}

        % dashed outer container
        \node[draw=white, fill=white, dashed, rounded corners,
              minimum width=0.75cm, minimum height=1.65cm] (AV_Solo) at (0,0) {};

        % task boxes
        \node[draw, rectangle, fill=soloFill, rounded corners,
              minimum width=0.85cm, minimum height=0.5cm] (TSR) at (0, 0.6)
              {\tiny \vlm$^{\text{TSR}}_{\text{Solo}}$};
        \node[draw, rectangle, fill=soloFill, rounded corners,
              minimum width=0.85cm, minimum height=0.5cm] (ALC) at (0, 0.0)
              {\tiny \vlm$^{\text{ALC}}_{\text{Solo}}$};
        \node[draw, rectangle, fill=soloFill, rounded corners,
              minimum width=0.85cm, minimum height=0.5cm] (VD)  at (0,-0.6)
              {\tiny \vlm$^{\text{VD}}_{\text{Solo}}$};

        % TSR arrows
        \draw[->] ($(TSR.north)+(-0.35,0.5)$) -- ($(TSR.north)+(-0.35,0)$)
          node[pos=-0.05,left] {\scriptsize Prompt\textsubscript{{\tiny TSR}}};
        \draw[->] ($(TSR.north)+(0.35,0.5)$) -- ($(TSR.north)+(0.35,0)$)
          node[pos=-0.05,right] {\scriptsize Data\textsubscript{{\tiny TSR}}};
        \draw[->] (TSR.east) -- ++(0.65,0)
          node[below,xshift=5pt] {\scriptsize Output\textsubscript{{\tiny TSR}}};

        % ALC arrows (longer horizontally)
        \draw[<-,shorten >=1pt] ($(ALC.west)+(0, 0.175)$) -- ++(-0.80,0)
          node[above,xshift=-3pt] {\scriptsize Prompt\textsubscript{{\tiny ALC}}};
        \draw[<-,shorten >=1pt] ($(ALC.west)+(0,-0.175)$) -- ++(-0.80,0)
          node[below] {\scriptsize Data\textsubscript{{\tiny ALC}}};
        \draw[->,shorten >=1pt] (ALC.east) -- ++(0.80,0)
          node[below,xshift=4pt] {\scriptsize Output\textsubscript{{\tiny ALC}}};

        % VD arrows
        \draw[->] ($(VD.south)+(-0.35,-0.5)$) -- ($(VD.south)+(-0.35,0)$)
          node[pos=0.6,left,yshift=-4pt] {\scriptsize Prompt\textsubscript{{\tiny VD}}};
        \draw[->] ($(VD.south)+(0.35,-0.5)$) -- ($(VD.south)+(0.35,0)$)
          node[pos=0.6,right,yshift=-4pt] {\scriptsize Data\textsubscript{{\tiny VD}}};
        \draw[->] (VD.east) -- ++(0.615,0)
          node[below,xshift=5pt] {\scriptsize Output\textsubscript{{\tiny VD}}};
      \end{tikzpicture}

      % one solid line ABOVE the subcaption:
      \vspace{2pt}\noindent\rule{\linewidth}{0.5pt}
    \end{minipage}%
  }\hfill
  %==================== Tandem Mode =====================
  \subfloat[\,Tandem Mode\label{fig:Tandem_Defense_Design}]{%
    \begin{minipage}[t]{0.40\linewidth}
      \centering
      \begin{tikzpicture}[x=1cm,y=1cm,>=Latex]
        \path[use as bounding box] (-1.25,-1.25) rectangle (1.25,1.25);
        \definecolor{tandembox}{RGB}{244,244,244}

        % outer container
        \node[draw=white, fill=white, dashed, rounded corners,
              minimum width=0.875cm, minimum height=1.65cm] (AV_Tandem) at (0,0) {};

        % tandem box
        \node[draw, rectangle, fill=tandembox, rounded corners,
              minimum width=0.85cm, minimum height=0.5cm] (Tandem) at (0,0)
              {\scriptsize \vlm\textsubscript{{\tiny Tandem}}};

        % prompt (left)
        \draw[<-] ($(Tandem.west)+(0,-0.02)$) -- ($(Tandem.west)+(-0.6,-0.02)$)
          node[above,xshift=-4pt] {\scriptsize Prompt\textsubscript{{\tiny all}}};

        % three data inputs from top
        \draw[->] ($(Tandem.north)+(-0.60,0.75)$) -- ($(Tandem.north)+(-0.60,0)$)
          node[left,yshift=15pt,xshift=-1pt] {\scriptsize Data\textsubscript{{\tiny TSR}}};
        \draw[->] ($(Tandem.north)+(0,0.75)$)    -- ($(Tandem.north)+(0,0)$)
          node[above,yshift=20pt] {\scriptsize Data\textsubscript{{\tiny ALC}}};
        \draw[->] ($(Tandem.north)+(0.6,0.75)$)  -- ($(Tandem.north)+(0.6,0)$)
          node[right,yshift=15pt,xshift=-1pt] {\scriptsize Data\textsubscript{{\tiny VD}}};

        % output downward
        \draw[-] (Tandem.east) -- ++(0.2,0) coordinate (outputnode);
        \draw[->] (outputnode) -- ([yshift=-1.0cm]outputnode)
          node[left] {\scriptsize Output\textsubscript{{\tiny all}}};
      \end{tikzpicture}

      % one solid line ABOVE the subcaption:
      \vspace{2pt}\noindent\rule{\linewidth}{0.5pt}
    \end{minipage}%
  }

  \caption{Comparison of Solo and Tandem Modes.}
  \label{fig:Defense_Methods}
\end{figure}

\noindent \textbf{Solo vs. Tandem Design Comparison.}
Deploying \vlms to enhance robustness against AEs in AVs raises important considerations for their implementation. Given the diverse range of tasks that AVs must perform, it is crucial to determine the best strategy for utilizing them. One approach involves using a separate \vlm for each specific AV task to improve robustness within each module, ensuring specialized and precise detection capabilities. This design, referred to as \textit{Solo Mode}, is illustrated in \autoref{fig:Solo_Defense_Design}, where individual \vlms are dedicated to tasks such as TSR, ALC, and VD. The formal representation of this design is: 

\begin{equation}\label{eq:solo_eq}
\textit{Output}_i = \vlm^{i}_{\textit{solo}}(\textit{Data}_i, \textit{Prompt}_i)
\end{equation}

\noindent
where \( i \in \{\textit{TSR}, \textit{ALC}, \textit{VD}\} \). Here, \( \textit{Data}_i \) denotes the dataset specific to each perception task \( T_i \), and \( \textit{Prompt}_i \) is the prompt customized to fine-tune \(\vlm^{i}_{\textit{solo}}\) for that particular task, ensuring task-specific optimization.

Alternatively, a single \vlm can handle all AV tasks using a tandem approach, providing a unified method to improve robustness across multiple modules, as depicted in \autoref{fig:Tandem_Defense_Design}. In this design, multiple image query pairs, each corresponding to a different task, are combined into a single input: the images are concatenated using a separator and the queries are merged in the same order. The model processes this structured batch simultaneously within one forward pass, extracting task-specific outputs for each image-query pair independently while treating the collection as a cohesive input during execution:

{\setlength{\abovedisplayskip}{1pt}%
 \setlength{\belowdisplayskip}{1pt}%
 \setlength{\abovedisplayshortskip}{0pt}%
 \setlength{\belowdisplayshortskip}{0pt}%
 \begin{equation}\label{eq:tandem_eq}
   \smash{\resizebox{\linewidth}{!}{$
     \{ \textit{Output}_{\textit{TSR}}, \textit{Output}_{\textit{ALC}}, \textit{Output}_{\textit{VD}} \}
     = \vlm_{\textit{tandem}}(\textit{Data}_{\textit{all}}, \textit{Prompt}_{\textit{all}})
   $}}
 \end{equation}%
}

\noindent
where \( \textit{Data}_{\textit{all}} \) represents the combined dataset of all tasks, and \( \textit{Prompt}_{\textit{all}} \) includes the concatenated prompts corresponding to each task.
Moreover, a single \vlm could reduce memory and computational overhead, which is critical in the resource-constrained environment of AVs. In such systems, optimizing both efficiency and memory is vital, making a unified \vlm a more practical solution. 

%%%%%%%%%%%%%%%%%%%%%%%%
\begin{table}[t]
\caption{Range of zero-shot performance (\%) across these six VLMs for each AV perception task. Results reflect the lowest and highest values among all models before fine-tuning.}
\centering
\footnotesize
\renewcommand{\arraystretch}{1.2}
\resizebox{\columnwidth}{!}{%
\begin{tabular}{lcccc}
\hline
\textbf{Task} & \textbf{Accuracy (\%)} & \textbf{F1-Score (\%)} & \textbf{Precision (\%)} & \textbf{Recall (\%)} \\ \hline
\textbf{TSR} & 1.19–31.24 & 0.95–29.80 & 1.15–29.24 & 0.48–30.31 \\
\textbf{ALC} & 20.71–50.91 & 24.60–51.18 & 28.38–51.44 & 21.39–50.91 \\
\textbf{VD}  & 60.64–92.06 & 64.01–92.01 & 71.43–94.38 & 50.46–92.06 \\
\hline
\end{tabular}
}
\label{table:Pre-Fine-Tuning-av-solo}
\end{table}

Integrating \vlms with AV perception systems offers the potential to strengthen resilience against AEs in real-time. Currently, AV perception systems face significant limitations in mitigating attacks, which can lead to dangerous misinterpretations of sensor data. As Cao~\textit{et al.}~\cite{cao2019adversarial} demonstrated, despite efforts to enhance the security of AV perception, significant vulnerabilities persist and traditional detection mechanisms often fail to mitigate these threats, leading to potentially dangerous consequences for AV decision-making and safety. By integrating \vlm in the end-to-end AV stack, AVs can benefit from its ability to process data, thereby working in parallel with perception tasks to enhance robustness and support accurate decision-making. The outputs of the perception system and \vlm can then be used by the downstream \textit{planning} and \textit{control} modules to act accordingly in the presence of AEs. This context is depicted in~\autoref{fig:integration_AV_VLM}.

It is crucial to evaluate the latency of \vlm~integration to ensure it can enhance robustness against AEs in real-time without affecting the efficiency of the perception system. AV vision algorithms can process images at a rate of 2-4 frames per second (fps), which corresponds to approximately 250-500 ms per image \cite{blachut2022automotive}. %As a result, \vlm~should process inputs within the required real-time constraints to support robust perception.

\section{Experimental Evaluation}
\label{sec:5-evaluation}

\subsection{Experimental Setup}

To evaluate the efficacy of \vlms, the focus was on three critical tasks: Traffic Sign Recognition (TSR), Automated Lane Centering (ALC) and Vehicle Detection (VD). For the former, the German Traffic Sign Recognition Benchmark (GTSRB) was utilized, which includes images captured under various conditions such as different lighting and distances. Each picture from this dataset, which has 42 classes, was sent with the prompt \textit{"Identify this traffic sign."} to the respective \vlm. For the ALC task, a dataset was generated using the CARLA simulator, which includes 6,000 training and 2,000 testing images. These images, taken from the driver’s perspective under various weather conditions and times of day, are classified into three categories indicating the next move: \textit{Straight}, \textit{Left}, and \textit{Right}. The prompt used for this task was, \textit{"As a car driver, at which direction should you turn the steering wheel?"}
%\mert{Maybe you want to include a table with the prompts you used as an overview?}

For VD, CARLA was also used to create 7,000 training and 3,000 testing images. Captured from different viewpoints, under diverse weather conditions, at various times of day, and from different distances, these images are categorized into \textit{Car presence} and \textit{Car absence}. If a car is present, the bounding box coordinates are provided with $x$ and $y$ representing the center, and $H$ and $W$ indicating the height and width of the box. The prompt used for this task was, \textit{"If a car is detected, provide the center coordinates and the dimensions of the bounding box for the car"}. 

The evaluation began with assessing the VLMs' performance on zero-shot tasks across the test datasets. Subsequently, these VLMs were fine-tuned on AV-specific training data --- yielding \vlms --- with their performance re-evaluated on the same test datasets to ensure comparability. The objective is to use a \vlm to enhance the ALC, TSR, and VD modules within the AV perception system against adversarial attacks. In the next step, AEs were generated from the same test datasets to assess robustness. 

Although the fine-tuned models were trained on AV-specific data, the AEs remained \textbf{\textit{unseen}} during training, allowing a reliable evaluation of each \vlm's resilience to attacks on TSR, ALC, and VD. To achieve this, three types of black-box attacks were implemented. The first type of attack involves adversarial manipulation of traffic signs. In the study by Zhong \textit{et al.}~\cite{zhong2022shadows}, shadows are utilized to conduct  attacks on TSR algorithms, as shown in figure~\ref{fig:shadow_AE}. 

This method employs shadows as a non-invasive mechanism to create physical AEs. By optimizing shadow properties such as shape and opacity using a differentiable renderer, the technique manipulates images under black-box conditions to induce misclassifications. It achieves a success rate of 90.47\% on the GTSRB dataset, demonstrating its effectiveness and highlighting the vulnerabilities of current detection systems to such subtle manipulations.

%##########################################################%###################################################
\begin{table}[t]
\centering
\caption{Performance of DNN models on benign (B) and unseen adversarial (A). Diff is the accuracy/F1 drop due to attacks.}
\footnotesize
\setlength{\tabcolsep}{3pt}
\renewcommand{\arraystretch}{0.95}
\begin{tabular}{lllccc}
\toprule
\textbf{Task} & \textbf{Model} & \textbf{Metric} & \textbf{A} & \textbf{B} & \textbf{Diff} \\
\midrule
\multirow{4}{*}{\textbf{TSR}} 
  & \multirow{2}{*}{YOLOv5-cls} 
    & Acc. & 55.82\% & 94.88\% & \cellcolor{LightRed}-39.06\% \\
  & 
    & F1-Score & 58.10\% & 93.44\% & \cellcolor{LightRed}-35.34\% \\
\cmidrule(lr){2-6}
  & \multirow{2}{*}{ViT-small} 
    & Acc. & 23.12\% & 96.20\% & \cellcolor{LightRed}-73.08\% \\
  & 
    & F1-Score & 23.12\% & 96.36\% & \cellcolor{LightRed}-73.24\% \\
\midrule
\multirow{2}{*}{\textbf{ALC}} 
  & \multirow{2}{*}{CLRerNet} 
    & Acc. & 50.51\% & 90.91\% & \cellcolor{LightRed}-40.40\% \\
  & 
    & F1-Score & 44.86\% & 90.83\% & \cellcolor{LightRed}-45.97\% \\
\midrule
\multirow{2}{*}{\textbf{VD}} 
  & \multirow{2}{*}{YOLOv5-dt} 
    & Acc. & 62.27\% & 97.01\% & \cellcolor{LightRed}-34.74\% \\
  & 
    & F1-Score & 57.44\% & 96.60\% & \cellcolor{LightRed}-39.16\% \\
\bottomrule
\end{tabular}
\label{table:dnn_on_benign_AE}
\end{table}

The second type of attack targets the ALC mechanism of AVs. In the study by Sato \textit{et al.}~\cite{sato2021dirty}, the Dirty Road Patch (DRP) attack framework specifically targets ALC systems in AVs, exploiting vulnerabilities in deep learning-based lane detection. This method employs an optimization-based approach to systematically generate these patches, as illustrated in figure~\ref{fig:DRP}, considering real-world conditions such as lighting and camera angles to ensure effectiveness across different environmental scenarios. The optimized DRPs cause the AV to make incorrect steering decisions, which were demonstrated to be highly successful in real-world driving scenarios with a success rate exceeding 97.5\%

Figure~\ref{fig:RAUCA} shows the third type of attack which focuses on the adversarial camouflage of vehicles. Zhou \textit{et al.}~\cite{zhou2024rauca} propose a physical adversarial attack known as the Robust and Accurate UV-map-based Camouflage Attack (RAUCA) to deceive VD algorithms such as YOLOv3~\cite{redmon2018yolov3}. It employs a technique utilizing a differentiable neural renderer, which allows for the optimization of adversarial camouflages through gradient back-propagation, enhancing both the robustness and precision of the attacks under varying environmental conditions. Their method achieved an attack success rate of 97.48\% on the target detection models, demonstrating the significant vulnerability of these systems to such sophisticated camouflage attacks.

\begin{table*}[t]
\centering
\tiny
\caption{comparison of \vlm\ performance: Benign vs. AEs. Green shows improvement of Tandem over Solo.}
\label{table:merged-av-vs-ae-acc-f1}
\begin{adjustbox}{width=\textwidth}
\setlength{\tabcolsep}{5pt}
\begin{tabular}{p{.3cm}p{1.5cm}|ccc|ccc||ccc|ccc}
    \hline
    \multicolumn{2}{c|}{} & \multicolumn{6}{c||}{\textbf{Benign}} & \multicolumn{6}{c}{\textbf{AEs}} \\
    \cline{3-14}
    \multicolumn{2}{c|}{} & \multicolumn{3}{c|}{\textbf{Accuracy}} & \multicolumn{3}{c||}{\textbf{F1-Score}}
                           & \multicolumn{3}{c|}{\textbf{Accuracy}} & \multicolumn{3}{c}{\textbf{F1-Score}} \\
    \cline{3-14}
    \textbf{Task} & \textbf{Model} & Solo & Tandem & Diff & Solo & Tandem & Diff & Solo & Tandem & Diff & Solo & Tandem & Diff \\
    \hline
\multirow{6}{*}{\textbf{TSR}}
    & \texttt{LLaVA-7B}  & 95.93\% & 97.38\% & \cellcolor{LightGreen}1.45\%
                         & 94.62\% & 91.22\% & \cellcolor{LightRed}-3.40\%
                         & 80.12\% & 86.51\% & \cellcolor{LightGreen}6.39\%
                         & 86.44\% & 85.89\% & \cellcolor{LightRed}-0.55\% \\
    & \texttt{LLaVA-13B} & 97.15\% & 98.14\% & \cellcolor{LightGreen}0.99\%
                         & 96.09\% & 93.21\% & \cellcolor{LightRed}-2.88\%
                         & 86.53\% & 89.01\% & \cellcolor{LightGreen}2.48\%
                         & 86.08\% & 88.95\% & \cellcolor{LightGreen}2.87\% \\
    & \texttt{MoE-LLaVA} & 95.24\% & 96.91\% & \cellcolor{LightGreen}1.67\%
                         & 94.73\% & 95.78\% & \cellcolor{LightGreen}1.05\%
                         & 80.03\% & 81.52\% & \cellcolor{LightGreen}1.49\%
                         & 80.75\% & 82.34\% & \cellcolor{LightGreen}1.59\% \\
    & \texttt{MobileVLM} & 88.19\% & 90.34\% & \cellcolor{LightGreen}2.15\%
                         & 87.68\% & 90.18\% & \cellcolor{LightGreen}2.50\%
                         & 79.57\% & 79.13\% & \cellcolor{LightRed}-0.44\%
                         & 77.55\% & 79.88\% & \cellcolor{LightGreen}2.33\% \\
    & \texttt{Qwen-VL}   & 95.80\% & 94.13\% & \cellcolor{LightRed}-1.67\%
                         & 95.78\% & 94.27\% & \cellcolor{LightRed}-1.51\%
                         & 82.83\% & 81.66\% & \cellcolor{LightRed}-1.17\%
                         & 83.16\% & 83.43\% & \cellcolor{LightGreen}0.27\% \\
    & \texttt{NVILA-8B}  & 99.04\% & 99.14\% & \cellcolor{LightGreen}0.1\%
                         & 99.10\% & 99.20\% & \cellcolor{LightGreen}0.1\%
                         & 86.04\% & 91.12\% & \cellcolor{LightGreen}5.08\%
                         & 87.15\% & 91.10\% & \cellcolor{LightGreen}3.95\% \\
    \cline{2-14}
\multirow{6}{*}{\textbf{ALC}}
    & \texttt{LLaVA-7B}  & 98.11\% & 95.41\% & \cellcolor{LightRed}-2.70\%
                         & 98.05\% & 95.24\% & \cellcolor{LightRed}-2.81\%
                         & 86.07\% & 84.83\% & \cellcolor{LightRed}-1.24\%
                         & 87.01\% & 85.38\% & \cellcolor{LightRed}-1.63\% \\
    & \texttt{LLaVA-13B} & 99.30\% & 98.31\% & \cellcolor{LightRed}-0.99\%
                         & 99.46\% & 97.85\% & \cellcolor{LightRed}-1.61\%
                         & 90.05\% & 86.69\% & \cellcolor{LightRed}-3.36\%
                         & 90.69\% & 87.38\% & \cellcolor{LightRed}-3.31\% \\
    & \texttt{MoE-LLaVA} & 96.86\% & 92.53\% & \cellcolor{LightRed}-4.33\%
                         & 96.23\% & 91.83\% & \cellcolor{LightRed}-4.40\%
                         & 82.54\% & 82.26\% & \cellcolor{LightRed}-0.28\%
                         & 83.27\% & 83.06\% & \cellcolor{LightRed}-0.21\% \\
    & \texttt{MobileVLM} & 84.41\% & 86.61\% & \cellcolor{LightGreen}2.20\%
                         & 87.68\% & 89.53\% & \cellcolor{LightGreen}1.85\%
                         & 78.21\% & 78.23\% & \cellcolor{LightGreen}0.02\%
                         & 80.92\% & 81.21\% & \cellcolor{LightGreen}0.29\% \\
    & \texttt{Qwen-VL}   & 93.75\% & 93.91\% & \cellcolor{LightGreen}0.16\%
                         & 93.83\% & 93.76\% & \cellcolor{LightRed}-0.07\%
                         & 88.81\% & 86.57\% & \cellcolor{LightRed}-2.24\%
                         & 89.48\% & 85.48\% & \cellcolor{LightRed}-4\% \\
    & \texttt{NVILA-8B}  & 99.51\% & 99.41\% & \cellcolor{LightRed}-0.1\%
                         & 99.45\% & 99.52\% & \cellcolor{LightGreen}0.07\%
                         & 99.25\% & 99.14\% & \cellcolor{LightRed}-0.11\%
                         & 99.13\% & 99.10\% & \cellcolor{LightRed}-0.03\% \\
    \cline{2-14}
\multirow{6}{*}{\textbf{VD}}
    & \texttt{LLaVA-7B}  & 95.84\% & 91.90\% & \cellcolor{LightRed}-3.94\%
                         & 94.56\% & 92.82\% & \cellcolor{LightRed}-1.74\%
                         & 89.23\% & 88.02\% & \cellcolor{LightRed}-1.21\%
                         & 87.84\% & 86.59\% & \cellcolor{LightRed}-1.25\% \\
    & \texttt{LLaVA-13B} & 97.05\% & 98.96\% & \cellcolor{LightGreen}1.91\%
                         & 96.76\% & 95.97\% & \cellcolor{LightRed}-0.79\%
                         & 91.82\% & 90.09\% & \cellcolor{LightRed}-1.73\%
                         & 90.72\% & 90.41\% & \cellcolor{LightRed}-0.31\% \\
    & \texttt{MoE-LLaVA} & 95.52\% & 93.42\% & \cellcolor{LightRed}-2.10\%
                         & 94.78\% & 92.30\% & \cellcolor{LightRed}-2.48\%
                         & 88.46\% & 86.54\% & \cellcolor{LightRed}-1.92\%
                         & 87.16\% & 86.21\% & \cellcolor{LightRed}-0.95\% \\
    & \texttt{MobileVLM} & 86.61\% & 88.01\% & \cellcolor{LightGreen}1.40\%
                         & 88.42\% & 89.38\% & \cellcolor{LightGreen}0.96\%
                         & 80.13\% & 80.15\% & \cellcolor{LightGreen}0.02\%
                         & 82.88\% & 82.97\% & \cellcolor{LightGreen}0.09\% \\
    & \texttt{Qwen-VL}   & 96.02\% & 95.01\% & \cellcolor{LightRed}-1.01\%
                         & 96.55\% & 96.17\% & \cellcolor{LightRed}-0.38\%
                         & 89.05\% & 89.06\% & \cellcolor{LightGreen}0.01\%
                         & 90.15\% & 90.57\% & \cellcolor{LightGreen}0.42\% \\
    & \texttt{NVILA-8B}  & 99.93\% & 99.95\% & \cellcolor{LightGreen}0.02\%
                         & 99.83\% & 99.84\% & \cellcolor{LightGreen}0.01\%
                         & 99.81\% & 99.83\% & \cellcolor{LightGreen}0.02\%
                         & 99.82\% & 99.83\% & \cellcolor{LightGreen}0.01\% \\
    \hline
\end{tabular}
\end{adjustbox}
\end{table*}

%##########################################################%##########################################################%######################################################

\begin{figure}[b]
  \centering
  \small
  \begin{tikzpicture}
    \begin{axis}[
      width=\columnwidth,          % span one column
      height=4cm,
      ybar,
      bar width=0.15cm,            % slimmer bars → more gap
      enlarge x limits=0.25,       % 25% padding left/right
      ymin=0, ymax=1.1,
      axis line style={draw=none},
      ymajorgrids,
      major grid style={draw=gray!40, dashed},
      ytick={0,0.2,0.4,0.6,0.8,1.0},
      ylabel={IoU Value},
      % list coords in the same order as xticklabels below
      symbolic x coords={
        LLaVA-7B,
        LLaVA-13B,
        MoE-LLaVA,
        MobileVLM,
        Qwen-VL,
        NVILA-8B
      },
      xtick=data,
      xticklabels={
        {\shortstack[c]{LLaVA\\-7B}},
        {\shortstack[c]{LLaVA\\-13B}},
        {\shortstack[c]{MoE\\-LLaVA}},
        {\shortstack[c]{Mobile\\VLM}},
        {\shortstack[c]{Qwen\\-7B}},
        {\shortstack[c]{NVILA\\-8B}}
      },
      xticklabel style={
        font=\tiny,
        align=center
      },
      yticklabel style={font=\tiny},
      ylabel style={font=\scriptsize},
      legend style={
        at={(0.5,1.1)},
        anchor=south,
        legend columns=3,
        font=\scriptsize
      },
      xtick style={draw=none},
    ]
      \addplot+[ybar, fill=none, draw=red,
                pattern=dots, pattern color=red]
        coordinates {
          (LLaVA-7B,0.041)   (LLaVA-13B,0.063)
          (MoE-LLaVA,0.038)  (MobileVLM,0.013)
          (Qwen-VL,0.018)    (NVILA-8B,0.183)
        };
      \addplot+[ybar, fill=none, draw=teal,
                pattern=crosshatch, pattern color=teal]
        coordinates {
          (LLaVA-7B,0.964)   (LLaVA-13B,0.987)
          (MoE-LLaVA,0.946)  (MobileVLM,0.926)
          (Qwen-VL,0.9578)   (NVILA-8B,0.9768)
        };
      \addplot+[ybar, fill=none, draw=blue,
                pattern=north east lines, pattern color=blue]
        coordinates {
          (LLaVA-7B,0.955)   (LLaVA-13B,0.971)
          (MoE-LLaVA,0.938)  (MobileVLM,0.913)
          (Qwen-VL,0.8920)   (NVILA-8B,0.9801)
        };
      \legend{
        Pre-trained,
        Fine-Tuned (Solo),
        Fine-Tuned (Tandem)
      }
    \end{axis}
  \end{tikzpicture}
  \caption{IoU values before and after fine‐tuning.}
  \label{fig:IoU}
\end{figure}

\subsection{RQ1: Fine-Tuning Increases Detection Performance}

\autoref{table:Pre-Fine-Tuning-av-solo} shows VLMs' zero-shot performance on the test dataset, which performed poorly in ALC and TSR tasks, indicating difficulties in these specific AV applications. Although the models demonstrated decent performance in the VD task, this success may partly be due to their pre-training on large, diverse image datasets, which likely enhanced their general visual recognition capabilities. However, in the VD task, they struggled to detect vehicles in images taken at night, in the rain, or at long distances (highlighting their limitations under challenging environmental conditions despite strong general visual recognition).

To improve their performance, these models were fine-tuned using the same training datasets and prompts as before, and then their performance was re-evaluated with the same test dataset. \autoref{table:merged-av-vs-ae-acc-f1} shows notable accuracy improvements for all tasks after fine-tuning. For ALC, accuracy increased from 20.71\%–50.91\% to 86.61\%–99.51\%. For TSR, accuracy saw a substantial rise from 1.19\%–31.24\% to 90.34\%–99.14\%. In the VD task, accuracy improved from 60.64\%–92.06\% to 88.01\%–99.95%.

These findings suggest that although they initially underperformed on AV tasks, fine-tuning them with relevant datasets can lead to substantial performance improvements, showing their potential utility in AV applications. Intersection over Union (IoU) measures the overlap between the predicted output and the ground truth in tasks like object detection. A higher IoU indicates more accurate localization, making it a crucial metric for evaluating model performance in AV perception tasks. After fine-tuning, the IoU values improved significantly, with \texttt{LLaVA-7B} increasing from 0.041 to 0.964, \texttt{LLaVA-13B-LoRA} from 0.063 to 0.987, \texttt{MoE-LLaVA} from 0.038 to 0.946, \texttt{MobileVLM} from 0.013 to 0.926, \texttt{Qwen-VL} from 0.018 to 0.9578, and \texttt{NVILA-8B} from 0.183 to 0.9768, as shown in~\autoref{fig:IoU}.

\subsection{RQ2: \vlms Demonstrate Robustness under Attacks}

Adversarial attacks pose a serious threat to deep learning based AV perception systems. Prior works have shown that DNN models suffer significant performance degradation across core AV tasks, including TSR, ALC, and VD, as summarized in~\autoref{table:attack_success_rate}. For example, the GTSRB CNN model accuracy collapses to just 1.77\% under the Shadow Attack~\cite{zhong2022shadows}, and even adversarial training, one of the most prominent defense methods, only modestly improves accuracy to 25.57\%. Similarly, for ALC, OpenPilot ALC performance degrades dramatically to 2.50\% under the DRP Attack~\cite{sato2021dirty}, with established defense strategies such as JPEG compression~\cite{guo2017countering}, Gaussian noise addition~\cite{xie2017mitigating}, and autoencoder based denoising~\cite{gu2014towards} achieving negligible improvements (around 3\%). In VD, YOLOv3 experiences a drastic accuracy drop to 2.52\% under RAUCA Attack~\cite{zhou2024rauca}, with no effective defense method proposed.

Building on these findings, we conduct our own robustness evaluation. To provide a fair and rigorous comparison, we evaluated traditional task-specific DNNs—\texttt{YOLOv5-cls} and \texttt{ViT-small} (TSR), \texttt{CLRerNet} (ALC), and \texttt{YOLOv5-dt} (VD)—on benign and previously \textit{unseen} adversarial datasets (\autoref{table:dnn_on_benign_AE}). Our results confirm severe performance degradation, with accuracy reductions of 73.08\% and 39.06\% for TSR, 40.40\% for ALC, and 32.91\% for VD, clearly highlighting their vulnerability to novel attacks. In contrast, our evaluations of \vlms under the same unseen adversarial conditions reveal inherently superior robustness. Specifically, \vlms experienced substantially smaller accuracy drops: only 8.62\%–15.81\% for TSR, 4.94\%–14.32\% for ALC, and 5.23\%–7.06\% for VD, consistently maintaining high adversarial accuracy without additional defenses. %These results strongly support the potential of \vlms to enhance robustness and reliability of AV perception systems.

\subsection{RQ3: Tandem \vlms Provide Similar Performance at Lower Memory Footprint}

We evaluate whether one \textit{tandem} \vlm provides robustness across AV tasks comparable to separate \textit{solo} \vlms. To assess this, VLMs are fine-tuned for three tasks, TSR, ALC, and VD, simultaneously using the same prompts. After fine-tuning, both solo and tandem \vlms were then evaluated on each task to determine its performance on the same test data. \autoref{table:merged-av-vs-ae-acc-f1} presents the evaluation results for the tandem design in non-adversarial scenarios, where a single \vlm was trained to handle all three tasks simultaneously. The results reveal that the tandem design achieves high accuracy across perception tasks, often matching or surpassing the performance of solo models. This shows the effectiveness of the tandem design in maintaining robust performance across diverse tasks. \autoref{table:merged-av-vs-ae-acc-f1} further demonstrates the resilience of \vlms in AV tasks under adversarial conditions. \texttt{MobileVLM}, \texttt{MoE-LLaVA}, \texttt{LLaVA-7B}, \texttt{NVILA-8B}, \texttt{Qwen-VL}, and \texttt{LLaVA-13B-LoRA} allocated 6.03GB, 11.21GB, 13.56GB, 15.23GB, 16.58GB, and 26.15GB of storage, respectively. These results suggest that a single tandem \vlm can generalize well across multiple tasks, providing robust performance comparable to the solo design, which requires 3x more storage for separate models. The tandem design offers a significant advantage in efficiency by keeping similar performance while requiring much less storage.

\begin{table}[t]
\tiny
\caption{DNNs accuracy under attacks and defenses.}
\centering
{\fontsize{16}{18}\selectfont
\renewcommand{\arraystretch}{1.3} % increased row spacing
\resizebox{\columnwidth}{!}{
\begin{tabular}{llcccc}
\hline
\addlinespace[5pt]
\textbf{Task} & \textbf{Model} & \textbf{\shortstack{Attack\\Type}} & \textbf{\shortstack{Under-Attack\\Accuracy}} & \textbf{\shortstack{Defense\\Strategy}} & \textbf{\shortstack{Post-Defense\\Accuracy}} \\ \hline
\addlinespace[5pt]

\multirow{1}{*}{\textbf{TSR}} 
  & \makecell[l]{GTSRB-\\CNN} 
  & \makecell[c]{Shadow \cite{zhong2022shadows}}  
  & \makecell[c]{1.77\%} 
  & \makecell[c]{Adversarial Training~\cite{madry2017towards}}  
  & \makecell[c]{25.57\%} \\ \cline{1-6} \addlinespace[5pt]

\makecell[c]{\textbf{ALC}} 
  & \makecell[l]{OpenPilot-\\ALC} 
  & \makecell[c]{DRP-Attack \cite{sato2021dirty}} 
  & \makecell[c]{2.50\%} 
  & \makecell[c]{JPEG Compression~\cite{guo2017countering} \\ Bit-Depth Reduction~\cite{xu2017feature} \\ Gaussian Noise~\cite{xie2017mitigating} \\ Median Blurring~\cite{xu2017feature}}  
  & \makecell[c]{$\approx$3\% \\ (no effective \\ improvement) } \\

\addlinespace[5pt]
\hline
\addlinespace[5pt]

\multirow{1}{*}{\textbf{VD}} 
  & \makecell[l]{YOLOv3} 
  & \makecell[c]{RAUCA \cite{zhou2024rauca}}
  & \makecell[c]{2.52\%} 
  & \makecell[c]{No Defense Proposed}
  & \makecell[c]{N/A} \\

\addlinespace[5pt]
\hline
\end{tabular}
}
}
\label{table:attack_success_rate}
\end{table}

%##########################################################%##########################################################%################################

\section{Discussion}
\label{sec:5-Discussion}

AV perception systems typically target latencies below 100 milliseconds to meet real-time operational requirements, especially in high-speed driving contexts \cite{jin2023benchmarking}. In our experiments, the inference time $t_{\text{\vlm}}$ for \texttt{LLaVA-7B} was measured at 851~ms on an NVIDIA A100 GPU with 40~GB of VRAM, clearly exceeding the acceptable threshold $t_{PS}$ for AV deployment. In contrast, the release of \texttt{NVILA} in late 2024 marked a significant improvement, reducing $t_{\text{\vlm}}$ to just 80~ms. This 10$\times$ reduction highlights the rapid evolution of VLMs toward real-time readiness in AV perception pipelines. Despite this progress, deploying VLMs on embedded AV hardware remains challenging due to limitations in compute power and energy efficiency. High-performance models such as \texttt{LLaVA-7B} and \texttt{NVILA}, though effective on server-grade GPUs, often require substantial memory and parallel processing capabilities that are impractical for in-vehicle deployment. One potential solution to reduce hardware demands is quantization, a widely used model compression technique for LLMs that improves computational efficiency by converting high-precision data types to lower-precision formats~\cite{IBMQuantization}. 

This process significantly reduces memory usage and model size, making it more feasible to run VLMs on edge devices. However, it may also introduce quantization errors, which could degrade precision~\cite{lin2024awq}. We applied quantization to \texttt{LLaVA-7B}, expecting lower resource usage. Although initially expected to reduce latency, the quantization approach did not yield the desired outcome; instead, when tested on the NVIDIA A100 40~GB, it exhibited latencies ranging from 0.851~s to 2.158~s and 11.657~s at full precision, 8-bit, and 4-bit levels, respectively. Based on our preliminary analysis, we found that the main source of this issue could be due to an increase in demand stemming from the additional operations required to maintain accuracy in image processing tasks.

\section{Conclusion}
\label{sec:6-conclusion}

% We present \vlms as a novel approach to enhance the robustness of AV perception systems against adversarial attacks. We evaluate \textit{Solo Mode} and \textit{Tandem Mode}, and demonstrate that \vlms maintain high adversarial accuracy without adversarial training while reducing storage requirements and maintaining comparable performance. Experimental results show that task-specific DNNs suffer performance drops of 33\%--74\% under adversarial attacks, whereas \vlms have reductions of less than 8\% on average. These results suggest that \vlms provide a promising foundation for inherently robust AV perception, and motivate future work on extending them to additional perception tasks, integrating other sensor modalities, and deploying them efficiently on resource-constrained AV hardware.

We present \vlms as a novel approach to enhance the robustness of AV perception systems against adversarial attacks. We evaluate \textit{Solo Mode} and \textit{Tandem Mode}, and demonstrate that \vlms maintain high adversarial accuracy without adversarial training while reducing storage requirements and maintaining comparable performance. Experimental results show that task-specific DNNs suffer performance drops of 33\%--74\% under adversarial attacks, whereas \vlms have reductions of less than 8\% on average.

\section*{Acknowledgment}
We gratefully acknowledge the support provided by the U.S. Department of Transportation (DOT) through the National Center for Transportation Cybersecurity and Resiliency (TraCR) under Grant No. 69A3552344812-2027534 and 69A3552348317. 
This work has also been partially supported by NSF under grant CNS-2443252 and The BMW Group. 
This research is with support from Google.org and the Google Cloud Research Credits program for the Gemma Academic Program.
% The authors also appreciate the support from Google GCP Credit Award program.

% % ------------------ References -------------------------------------------------
% \bibliographystyle{ieeeconf} % ICRA expects IEEE style
% \bibliography{ref}

\bibliographystyle{IEEEtran}
\bibliography{ref}

\end{document}